\documentclass[conference]{IEEEtran}

\pdfoutput=1
\usepackage[utf8]{inputenc}

\usepackage[T1]{fontenc}
\usepackage{url}
\usepackage{flushend}
\usepackage{xspace}
\usepackage[skip=5pt]{caption}
\usepackage[inline]{enumitem}
\usepackage{ifthen}
\usepackage{balance}
\usepackage{hyperref}
\usepackage[numbers]{natbib}
\usepackage{flushend}
\usepackage{orcidlink}

\usepackage{microtype}
\usepackage{multirow}
\usepackage[]{siunitx}
\newcommand*\np[2][z]{
\ifx z#1%
$\num{#2}$%
\else%
$#2\ #1$
\fi\xspace%
}
\usepackage{fp}
\usepackage{xfp}

\usepackage{graphicx}
\usepackage{subcaption}
\usepackage{pifont}

\usepackage{booktabs}
\usepackage{multirow}
\usepackage{threeparttable}
\usepackage{xcolor}
\usepackage{array,makecell}
\usepackage{diagbox}
\usepackage{tabularx}

\usepackage{mathtools}
\usepackage{amsmath}

\usepackage{tcolorbox}
\usepackage{pgfplots}
\usepackage{pgfplotstable}

\usepackage{enumitem}
\usepackage[framemethod=TikZ]{mdframed}
\usepackage[scaled]{beramono}

\usepgfplotslibrary{statistics}
\usetikzlibrary{calc,patterns,fpu} 
\pgfplotsset{compat=1.14} 

\pgfplotsset{
    /pgf/declare function={
        Floor(\x) = round(\x-0.49);
    },
    show sum on top/.style={
        /pgfplots/scatter/@post marker code/.append code={%
            \path let \p1=($(normalized axis cs:%
                        \pgfkeysvalueof{/data point/x},%
                        \pgfkeysvalueof{/data point/y})%
                        -(normalized axis cs:\pgfkeysvalueof{/data point/x},0)$)
            in node[
                at={(normalized axis cs:%
                        \pgfkeysvalueof{/data point/x},%
                        \pgfkeysvalueof{/data point/y})%
                },
                anchor={-90*sign(\y1)},yshift={sign(\y1)*2pt}
            ]
            {\pgfmathprintnumber[fixed, precision=1]{\pgfkeysvalueof{/data point/y}}};
        },
    }
}

\definecolor{blau_1a}{RGB}{93,133,195}
\definecolor{blau_2a}{RGB}{0,156,218}
\definecolor{gruen_3a}{RGB}{80,182,149}
\definecolor{gruen_4a}{RGB}{175,204,80}
\definecolor{gruen_5a}{RGB}{221,223,72}
\definecolor{orange_6a}{RGB}{255,224,92}
\definecolor{orange_7a}{RGB}{248,186,60}
\definecolor{rot_8a}{RGB}{238,122,52}
\definecolor{rot_9a}{RGB}{233,80,62}
\definecolor{lila_10a}{RGB}{201,48,142}
\definecolor{lila_11a}{RGB}{128,69,151}

\definecolor{blau_1b}{RGB}{0,90,169}
\definecolor{blau_2b}{RGB}{0,131,204}
\definecolor{gruen_3b}{RGB}{0,157,129}
\definecolor{gruen_4b}{RGB}{153,192,0}
\definecolor{gruen_5b}{RGB}{201,212,0}
\definecolor{orange_6b}{RGB}{253,202,0}
\definecolor{orange_7b}{RGB}{245,163,0}
\definecolor{rot_8b}{RGB}{236,101,0}
\definecolor{rot_9b}{RGB}{230,0,26}
\definecolor{lila_10b}{RGB}{166,0,132}
\definecolor{lila_11b}{RGB}{114,16,133}

\newboolean{showcomments}
\setboolean{showcomments}{true}

\newcommand{\ShowAbsoluteNumber}[1]{%
\ifnum #1<10%
{\hspace*{0pt}#1}%
\else%
\ifnum #1<100%
{\hspace*{0pt}#1}%
\else%
\ifnum #1<1000%
{\hspace*{0pt}#1}%
\else%
{\numprint{#1}}%
\fi%
\fi%
\fi%
}

\newcommand{\ShowPercentage}[2]{%
\FPeval\percentage{round(#1/#2*100,0)}%
\FPeval\percentageOneDecimal{round(#1/#2*100,1)}%
\ifnum \percentage=0%
{\np[\%]{\FPprint{percentageOneDecimal}}}%
\else%
\ifnum \percentage<10%
{\np[\%]{\FPprint{percentageOneDecimal}}}%
\else%
{\np[\%]{\FPprint{percentageOneDecimal}}}%
\fi%
\fi%
\xspace
}
\newlength\BARSIZE  
\newcommand{\inlinechart}[2]{%
\FPeval{\BLACKBARSIZE}{#1/#2}\textcolor{black!80}{\rule{\BLACKBARSIZE\BARSIZE}{1.6ex}}%
\FPeval{\BLACKBARSIZE}{1 - (#1/#2)}\textcolor{black!10}{\rule{\BLACKBARSIZE\BARSIZE}{1.6ex}}%
}

\newcommand*\percent[3][v]{%
\ifx q#1%
    \np{#2}/\np{#3}(\ShowPercentage{#2}{#3})\else%
\ifx s#1%
    \ShowPercentage{#2}{#3}\else%
\ifx p#1%
    \np{#2}(\ShowPercentage{#2}{#3})\else%
\ifx c#1%
    \inlinechart{#2}{#3}%
\else%
    \np{#2}%
    \ifx r#1%
        /\np{#3}%
    \fi%
    \hspace*{0.5ex}(\ShowPercentage{#2}{#3}) %
    \inlinechart{#2}{#3}%
    \xspace
\fi\fi\fi\fi%
}

\makeatletter
\@ifclasswith{IEEEtran}{anonymous}{%

}{%

}
\makeatother

\definecolor{mygray}{RGB}{240,240,240}

\ifthenelse{\boolean{showcomments}}
 { }
        { }

\definecolor{eminence}{RGB}{108,48,130}
\definecolor{weborange}{RGB}{255,165,0}
\definecolor{frenchplum}{RGB}{129,20,82}
\definecolor{darkgreen}{RGB}{10, 92, 10}

\mdfdefinestyle{mpdframe}{
    nobreak                     =true,
    backgroundcolor             =black!3,
    frametitlebackgroundcolor   =black!15,
    frametitlerule              =false,
    roundcorner                 =5pt,
    innermargin                 =0.3cm,
    innerleftmargin             =0.3cm,
    innerrightmargin            =0.3cm,
    innertopmargin              =0.3cm,
    innerbottommargin           =0.3cm,
    shadowsize                  =1pt
}

\begin{document}

\title{Energy Aware Development of Neuromorphic Implantables: From Metrics to Action}

\author{\IEEEauthorblockN{Enrique Barba Roque\orcidlink{0000-0002-7018-498X}}
\IEEEauthorblockA{\textit{Delft University of Technology} \\
Delft, The Netherlands \\
enrique@ebarba.com}
\and
\IEEEauthorblockN{Luis Cruz\orcidlink{0000-0002-1615-355X}}
\IEEEauthorblockA{\textit{Delft University of Technology} \\
Delft, The Netherlands \\
L.Cruz@tudelft.nl}
}

\maketitle

\begin{abstract}
Spiking Neural Networks (SNNs) and neuromorphic computing present a promising alternative to traditional Artificial Neural Networks (ANNs) by significantly improving energy efficiency, particularly in edge and implantable devices. However, assessing the energy performance of SNN models remains a challenge due to the lack of standardized and actionable metrics and the difficulty of measuring energy consumption in experimental neuromorphic hardware. In this paper, we conduct a preliminary exploratory study of energy efficiency metrics proposed in the SNN benchmarking literature. We classify 13 commonly used metrics based on four key properties: Accessibility, Fidelity, Actionability, and Trend-Based analysis. Our findings indicate that while many existing metrics provide useful comparisons between architectures, they often lack practical insights for SNN developers. Notably, we identify a gap between accessible and high-fidelity metrics, limiting early-stage energy assessment. Additionally, we emphasize the lack of metrics that provide practitioners with actionable insights, making it difficult to guide energy-efficient SNN development. To address these challenges, we outline research directions for bridging accessibility and fidelity and finding new Actionable metrics for implantable neuromorphic devices, introducing more Trend-Based metrics, metrics that reflect changes in power requirements, battery-aware metrics, and improving energy-performance tradeoff assessments. The results from this paper pave the way for future research on enhancing energy metrics and their Actionability for SNNs.
\end{abstract}


\section{Introduction}



Neuromorphic computing and Spiking Neural Networks (SNNs) are an emerging alternative to traditional Von Neumann computing\footnote{Traditional computing architecture based on shared memory and sequential execution of instructions.} and Artificial Neural Networks that aim to improve energy efficiency while conserving high accuracy.
This paradigm redesigns Neural Networks to more closely mimic the behaviour of the brain, encoding information across time in binary spikes, reducing the complexity of computation, and increasing parallelization.
These SNNs work in conjunction with the specialized neuromorphic hardware to reduce energy usage.

Design and fabrication of neuromorphic hardware are still experimental and in early stages. SNN models can be built in Python using libraries built on top of PyTorch, and can be later deployed to neuromorphic hardware.
While these models can be run on traditional CPU and GPU hardware, this execution will not reflect the energy efficiency gains that can be obtained from the neuromorphic hardware.
However, having access to neuromorphic hardware for deploying and testing the efficiency of the model is rather difficult, given the experimental nature of its components.
Therefore, being able to benchmark and detect inefficiencies in SNNs before deploying them into hardware is a relevant problem in the community.

Different works have tried to define standard SNN benchmarks, and metrics to evaluate accuracy and energy efficiency of these models \cite{yik2025neurobenchframeworkbenchmarkingneuromorphic, analyticalsnn, hueber2024benchmarking}.
However, while some of these metrics might be useful for ranking and comparison among different SNN architectures, they are difficult to interpret by developers.

Hence, there is an important gap in helping ML practitioners use the right metrics that allow them to select the most energy-efficient models that fulfill the requirements of the project.
We anticipate that there is no one-size-fits-all metric for energy efficiency, as it is highly dependent on the use case. 
Some use cases might require extending the lifetime of the battery as much as possible, while other use cases might focus on saving electricity bills, and others might prioritize reducing carbon emissions. 
Hence, different stakeholders have different needs, and different use cases require different perspectives.

This is particularly important in the context of this study, focused on the Smart Edge Lab for Healthcare (SELF lab) at TU Delft, where we are developing medical implants for early epilepsy detection\footnote{Epilepsy is a neurological disease that manifests as a brain-wide phenomenon.}. Such a device brings a multi-disciplinary team together that works at different layers of the implant: model development, hardware controller, hardware design, patient intervention, etc. Thus, we need different stakeholders with different disciplinary backgrounds to be able to communicate and make decisions based on energy metrics.  

In this exploratory paper, we present an overview of the state-of-the-art metrics for energy efficiency proposed in SNN benchmarking literature. Given that each metric has its pros and cons, we classify them according to different perspectives: Accessibility, Fidelity, Actionability, and Trend-Based.
We find that most metrics do not provide actionable insights to developers, and propose some research directions for alternative metrics that can improve energy assessment in the early stages of SNN model development.


\section{Background}\label{sec:background}
This Section introduces the necessary technical background to understand this paper.
We explain the main differences between Artificial Neural Networks and Spiking Neural Networks, and how the latter are directly supported by a new paradigm of neuromorphic hardware design.

\subsection{ANNs vs SNNs}

Advancements in the field of artificial intelligence have been predominantly influenced by the application of Artificial Neural Networks (ANN). 
Essentially, ANN models are composed of layers of interconnected neurons where each neuron receives inputs -- typically represented as floating point numbers -- processes these inputs through multiply-accumulate operations (MAC operations) and sends them to subsequent neurons through weighted connections. 
Learning methods, normally based on gradient descent and backpropagation, are used to train weights and improve the accuracy of the model.

Although the idea is not new, most advancements in this technology come from the ability to increase model complexity with more processing power~\cite{Tazi2025UltraScale}, adding more layers and more complex neurons, increasing the total arithmetic operations and, therefore, increasing energy consumption.
This makes ANNs suboptimal for low-power scenarios, like wearable devices or edge computing.

To tackle this energy efficiency problem, Spiking Neural Networks have emerged as an alternative. 
Taking inspiration from the human brain, these neural networks encode information in the form of discrete binary spikes across time \cite{MAASS19971659} rather than in the form of real numbers.
One of the most popular neuron models is the Leaky Integrate and Fire (LIF), initially developed by Lapicque in 1907 \cite{lapicque} to explain neurons' behavior, and later implemented for SNNs.
In this model, neurons keep an internal state -- called membrane potential, referencing brain mechanisms -- which increases when receiving spikes and makes the neuron fire a spike after a certain threshold.
When combined with neuromorphic hardware, this translates to multiple advantages compared to traditional ANNs.

Mainly, this design makes SNN an event-driven model~\cite{snnReview}. 
The activity in the hardware and energy usage happens mostly during spikes, which are relatively sparse. 
While a neuron is idle and not receiving or sending spikes, energy consumption is minimal.

Additionally, neuromorphic hardware is designed with high parallelization in mind. Processing and memory are co-located, forming one neuron, instead of having a central CPU and memory.
Therefore, multiple neurons can be operating simultaneously, and, since the spikes are binary values, most operations are simple accumulations (AC operations), and the number of MAC operations, which are typically more expensive, is greatly reduced.

Currently, the development of SNN models takes place mainly in Python. There are multiple libraries built on top of the PyTorch framework \cite{pytorch}, such as SNNTorch \cite{snntorch} or SpikingJelly \cite{spikingJelly}.
Combined with neuromorphic hardware, these SNNs can provide similar accuracy to ANN while using up to 100x less energy \cite{recurrentSpiking}.

\subsection{Neuromorphic hardware}
To completely exploit the energy efficiency gains of SNNs, these need to be deployed in neuromorphic hardware. 
This kind of hardware physically implements the concept of spiking neurons to mimic neuron models like the previously mentioned LIF. 
In contrast to traditional Von Neumann architecture, this hardware usually includes collocated processing and memory.
This leads to efficient event-driven parallel computing, where neurons can independently compute their results based on their inputs.
Additionally, the architecture is easily scalable by chaining additional neurons or chips.
Neuromorphic hardware can also be fully digital, using traditional binary computation, or mixed analog-digital design.
For this kind of design, LIF neurons are implemented with a circuit that includes a capacitor that stores the neuron's potential, and a comparator that fires when this potential is over a threshold voltage.

An example of this is SpiNNaker \cite{spinnaker}, which is composed of 57,600 interconnected nodes, each with 18 of these processors and 128 MB of memory.
Intel is also working on neuromorphic computing with Loihi and Loihi 2 \cite{loihi}.

BrainScale \cite{goltz2021fast} is a hybrid analog implementation of this concept. Their paper also provides a comparison between some of the previously mentioned architectures and their own, as well as the efficiency of traditional ANN with GPU hardware. They show that neuromorphic computing can have gains of up to 101x compared to traditional ANNs.

In conclusion, it is hard to measure and compare the energy efficiency of SNNs when they are not deployed into neuromorphic hardware that aims to maximize this efficiency.
For traditional ANNs, there are mature tools like RAPL \cite{rapl} and MLPerf \cite{mlperf} to benchmark models.
While these tools can be used for SNNs run in traditional hardware, the measurements obtained will hardly reflect the energy improvements that would be obtained with neuromorphic hardware.

\section{Research Context}\label{sec:self}

\begin{figure*}
    \centering    \includegraphics[width=.9\linewidth]{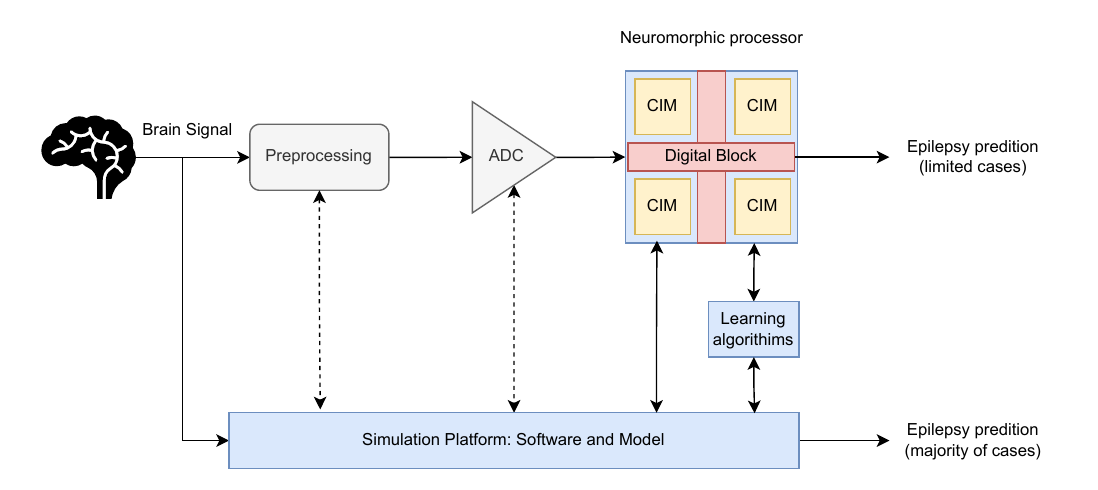}
    \caption{High-level overview of the SELF Lab objectives}
    \label{fig:self}
\end{figure*}

This paper stems from a real context in the SELF Lab where we need to deploy energy-efficient neural networks on implantable devices.
Concretely, the overarching project aims to develop an implantable neural device that runs an AI model for the early detection of epilepsy events, and early neurostimulation to prevent the episode. A high-level illustration of the project is presented in \autoref{fig:self}.

Therefore, the project brings together a multidisciplinary team with stakeholders who are involved across the entire device development pipeline. For example, part of the team works on hardware design and techniques to improve memristor hardware design \cite{rram}, while another part is working on learning algorithms for SNNs \cite{zhang2025nisnnanoniterativespikingneural}.
Ultimately, healthcare professionals and epilepsy patients also play an important role as they want to minimize any risks involved with implanting and using the smart edge device. 

In this context, energy consumption plays an important role: we want to maximize the lifetime of using the device without intervention. This means that the less energy the model spends, the less likely it is that the device needs to be recharged. Moreover, energy consumption is not only a question of battery lifetime, but also a question of capacity: power-hungry models will probably be physically impossible to run in such low-powered edge devices. Such a multidisciplinary context and ubiquitous scenario raise important challenges in the way we measure and communicate energy efficiency. Developers of AI models, hardware designers, healthcare professionals, and patients might have a different way of interpreting energy efficiency, and it is important that all stakeholders have a clear understanding of it so that optimal design decisions are made.

The metrics analysis presented in this paper is done through this lab lens. While some of our conclusions can be applied to the whole field of neuromorphic computing, we will focus especially on the value that these metrics bring to healthcare edge devices and how to facilitate the development of SNN for this specific application.


\section{Related Work}\label{sec:rel}

Despite the claims made by various papers about energy efficiency gains for SNNs, the standardization of benchmarking for SNNs' efficiency and accuracy still remains an open discussion in the community.
Many papers designed custom benchmarks to test their SNN design according to the task tackled by the model.
For example, SNN models that target computer vision will test their models on popular datasets like N-MNIST \cite{nmnist}, a neuromorphic adaptation of the traditional MNIST dataset for computer vision.
Then, they report energy consumption through different means, either by directly measuring on-chip usage -- mostly papers that focus on hardware design -- or by defining their own approximations based on performed operations or the complexity of the model.

The work by Ostrau et al. \cite{10.3389/fnins.2022.873935} introduces a set of benchmarks for open-loop tasks -- the system does not adapt its behaviour based on the output -- and compares the performance to the human brain based on individual costs of certain actions of the brain, and mapping those actions to neuromorphic chip operations.
However, these benchmarks still focus heavily on hardware rather than SNN model design, and lack formal definition or standardization.
Other tasks show a lack of benchmarking options, like closed-loop tasks -- the system acts based on predictions, e.g, detect an epilepsy attack and apply neurostimulation to avoid it -- as pointed by Milde et al. \cite{10.3389/fnins.2022.813555}

NeuroBench \cite{yik2025neurobenchframeworkbenchmarkingneuromorphic} is the first attempt at providing a standardized benchmarking framework, built on top of Pytorch and SNNTorch, in collaboration with a large part of the neuromorphic community. 
They provide multiple standardized datasets for different tasks, like computer vision or speech processing.

The framework facilitates the computation of a series of metrics for measuring both accuracy and estimating energy and resource usage of the model.
For representing energy usage of the model, the metrics provided focus on the number of synaptic operations (MAC and AC operations) and the number of membrane updates as main proxies to energy consumption.

However, later works agree that only measuring the number of operations is not enough to accurately estimate energy usage, and other metrics should be considered.
Hueber et al. \cite{hueber2024benchmarking} extend this benchmarking for brain-computer interfaces, using NeuroBench but considering memory footprint and number of memory accesses as additional metrics for estimating power consumption.
However, they derive the number of memory accesses from the number of operations computed by NeuroBench, estimating three loads and one store for each MAC operation, and two loads and one store for each AC operation.

On the other hand, Lemaire et al. \cite{analyticalsnn} also consider the MAC, AC, and memory accesses metrics, but propose an analytical estimation rather than runtime measurements, derived from the architecture of the model. 
They provide a series of equations to compute total operations and accesses for each layer of the SNN model and estimate energy consumption by assigning an energy cost to each type of operation and access, based on previous literature.

\section{Methodology}\label{sec:methodology}

\begin{figure*}
    \centering    \includegraphics[width=0.9\linewidth]{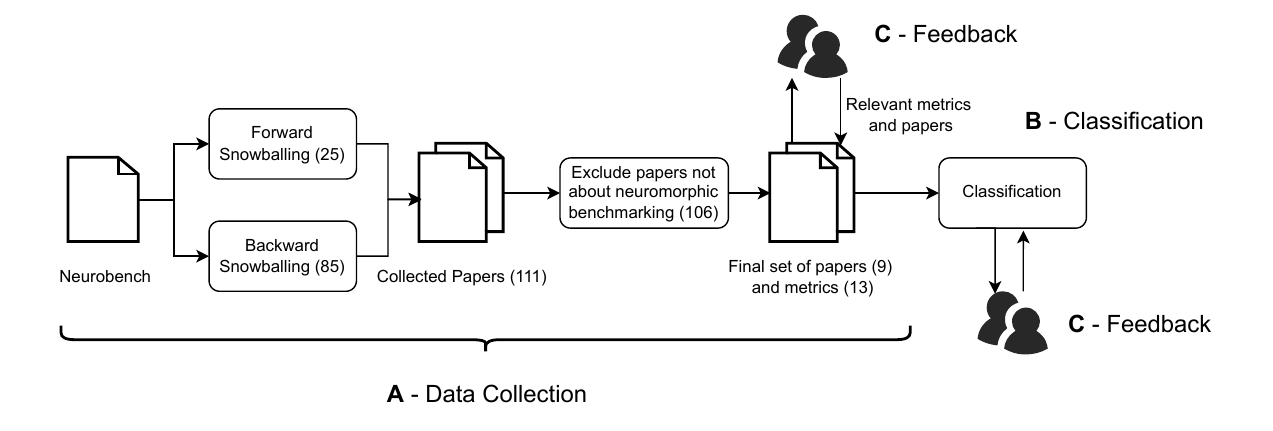}
    \caption{Methodology followed for finding and reviewing the metrics}
    \label{fig:methodology}
\end{figure*}

In this Section, we introduced the methodology followed for our review.
\autoref{fig:methodology} shows an overview of this methodology, which consists of three major steps.
First, we collect relevant papers on SNN benchmarking and energy metrics by applying snowballing from the NeuroBench paper.
Secondly, we classify the metrics according to 4 properties: Accessibility, Fidelity, Actionability, and Trend-Based.
Finally, we incorporate feedback from practitioners from the SELF Lab, who provide additional metrics and papers.

\subsection{Data collection}
While research on neuromorphic computing and SNNs has grown considerably in the last decade, works that focus on benchmarking energy efficiency are still relatively niche \cite{pritchard2023bibliometricreviewneuromorphiccomputing, snnReview}.
Therefore, the benchmarking literature is still not extensive, and papers that focus on energy measurement and estimation in this field are rare.
For this reason, we adopt an informal literature review to explore existing energy metrics in this context.

As pointed out in \autoref{sec:rel}, the standard observed in research is to create custom benchmarks to measure accuracy and energy efficiency.
NeuroBench~\cite{yik2025neurobenchframeworkbenchmarkingneuromorphic} was the first attempt to standardize the energy efficiency benchmarks for SNNs. However, later works criticized and proposed new metrics and methodologies.

We start by analyzing the metrics proposed in NeuroBench, and collect related work by performing forward and backward snowballing in the original NeuroBench paper~\cite{yik2025neurobenchframeworkbenchmarkingneuromorphic}, manually checking citations in NeuroBench and papers that cite NeuroBench. We used Google Scholar to gather forward and backward references available as of January 2025. Due to the recent nature of this topic, we have also considered preprints that are not yet peer-reviewed.
From this first selection, we only kept papers that focus on benchmarking of SNNs and discussions on energy metrics.
Afterwards, we discussed with practitioners from the SELF Lab, who pointed to additional papers and metrics.
In total, we analyzed 111 papers and found 9 papers that presented energy metrics for neuromorphic AI solutions.

\subsection{Classification}
In this section, we introduce our classification of the different energy performance metrics used in neuromorphic literature. We will analyze this from the Software Engineering perspective. This means that we will study the metrics from the perspective of an SNN model developer. We consider that this professional has knowledge of the theoretical foundations of SNNs and how to develop a working model that performs a specific task. However, the developer mostly works on traditional hardware (e.g., a work laptop) and might not have extensive knowledge about neuromorphic hardware, as well as limited access to the hardware where the SNN will be deployed.

In total, we collect 13 energy metrics applicable to the implantable neuromorphic field. We analyze these metrics based on four properties that we consider relevant in our context: Accessibility, Fidelity, Actionability, and Trend-Based. We pinpoint these different properties below.

\subsubsection{Accessibility}
In this work, we define a metric as having \textit{Accessibility} if it can be easily obtained by an SNN developer while working on their model without the need to access hardware.
In the context of neuromorphic computing, this would be a metric that is computed based on the model execution and does not require access to neuromorphic hardware to be obtained. 

As highlighted previously, an SNN developer will probably have limited access to the target hardware and might require assistance from the hardware designers in case of more experimental projects.
Accessible metrics can help detect energy inefficiencies in the model earlier, which can be ironed out before deploying to hardware, saving time and resources.

\subsubsection{Fidelity}
We say an energy metric has high \textit{Fidelity} if it accurately reflects the real energy usage of the model in production.
A metric with a high fidelity would be on-chip power measurements, but they are also the least accessible.
Therefore, we also consider other approximations of real energy usage that have been validated by research and industry.
For example, during the development of a new hardware architecture, simulation software is typically used to build the chip and test energy consumption, among other functionalities. 
While these results might not be exactly those observed with real-world measurements once the hardware is built, it is considered a pretty accurate approximation.

On the other hand, metrics like the number of operations do not have high fidelity.
While the total energy consumption of an SNN is directly related to the number of operations performed, the real energy usage will heavily depend on the hardware running the model.


\subsubsection{Actionability}
We say a metric provides \textit{Actionability} if it helps an SNN developer to decide if they should work on improving the energy efficiency of the model under development.

According to the findings by Ram et al. \cite{actionable}, actionable metrics must be practical, contextual, and exhibit high data quality (defined as accurate, consistent, and credible).

In our context, an actionable metric is practical if it is easy to interpret for the developer, and ideally, it reflects approachable concepts, rather than theoretical and non-intuitive measurements.
For example, the total number of MAC operations might reflect total energy, but it can be difficult to gauge if optimizations are necessary based only on that number.


While many of the metrics found in the literature reflect energy efficiency with different degrees of accuracy, many of them are not actionable.

\subsubsection{Trend-based}
Trend-based metrics are those that can indicate improvement or degradation of energy performance based on previous measurements.
These are metrics that, while not actionable per se, can give a sense of how changes to the model might affect energy consumption during deployment.

An example of this metric would be the number of parameters or the number of synaptic operations metrics mentioned previously. 
A single measurement of the number of operations required for an inference does not relay information about final energy costs. However, if a developer finds out that this number increases after making changes to the model, it would be a fair conclusion

\subsection{Feedback from practitioners}
To mitigate author biases in the classification of energy metrics, we ask three members of the SELF Lab to review the classification and provide their own perspectives on the presented metrics.

The team at the SELF Lab is multidisciplinary, with practitioners who handle hardware design and SNN model design.
One member works on mixed analog-digital neuromorphic hardware design and architecture.
Another member specializes in embedded and implantable Machine Learning devices.
Finally, the third member works on SNN model and algorithm design.

We asked them to provide their opinions on the properties defined, and asked them to classify the metrics we collected using the properties blindly, without access to our classification.
With this, we confirmed our classification and updated some metrics' properties based on the posterior discussion.
The practitioners also provided some additional papers that proposed metrics to look into.

\section{Results}\label{sec:metrics}


Following our methodology, we collect 13 metrics that are used to assess energy efficiency in neuromorphic software systems.
\autoref{tab:metrics} outlines each metric, providing a short description, our classification based on Accessibility, High Fidelity, Actionability, and Trend-Based properties, and references to neuromorphic literature that use the respective metric.

The ``Reference'' column indicates the paper or papers that discuss the metric as a metric for energy consumption in SNNs.
When accompanied with a P, it indicates that the metric was proposed by the practitioners.

Below, we examine each property in detail, analyzing how the metrics contribute to representing and informing developers about energy efficiency.


%

\begin{table*}[ht]
\centering
\caption{Metrics table with descriptions and attributes.}
\label{tab:metrics}
\begin{tabularx}{\textwidth}{|lXccccc|}
\hline
\textbf{Metric}               & \textbf{Description} & \textbf{Accessibility} & \textbf{High Fidelity} & \textbf{Actionability} & \textbf{Trend-Based} & \textbf{Reference} \\ \hline
Parameters                    & Trainable and/or non-trainable parameters. More can mean more energy. & Yes & No  & No  & No  & \cite{yik2025neurobenchframeworkbenchmarkingneuromorphic} \\ \hline
Effective Synaptic Operations & MACs and ACs in the synaptic connections of the network. & Yes & No  & No  & Yes  & \cite{yik2025neurobenchframeworkbenchmarkingneuromorphic, analyticalsnn,10.3389/fnins.2022.873935}  \\ \hline
Membrane Updates & MACs and ACs operations performed to update the potential of the neurons. & Yes & No  & No  & Yes  & \cite{yik2025neurobenchframeworkbenchmarkingneuromorphic, analyticalsnn, 10.3389/fnins.2022.873935} \\ \hline
Activation Sparsity           & Density of 0s across the whole network for an inference or group of inferences. & Yes & No  & Yes & Yes & \cite{yik2025neurobenchframeworkbenchmarkingneuromorphic} \\ \hline
Memory Footprint              & Size of the model in memory. & Yes & No  & No  & No  & \cite{hueber2024benchmarking, yik2025neurobenchframeworkbenchmarkingneuromorphic} \\ \hline
Connection Sparsity           & Connections between layers. & Yes & No  & No  & No  & \cite{yik2025neurobenchframeworkbenchmarkingneuromorphic} \\ \hline
Memory Accesses               & Read and write operations in memory. & Yes & No  & No  & Yes  & \cite{hueber2024benchmarking, analyticalsnn} \\ \hline
Training Time  & Total time taken to train the model. & Yes & No & No & Yes & \cite{karilanova2024zero,KULKARNI2021145} \\ \hline
Energy per Inference     & Energy consumption of performing a single inference task. Used in SoC ML. & No & Yes  & No  & No  & \cite{ke2024neurobench,chen2024epoc} \\ \hline
Energy per Learning & Energy consumption of processing a training sample.    & No & Yes  & No  & No & \cite{chen2024epoc} \\ \hline
Energy Area FoM               & Considers power per channel, area, and sampling frequency. Relatively novel, not used. & No  & Yes  & No  & No  & \cite{energyArea} \\ \hline
Peak per Energy Consumption     & Energy consumed per system operation (SOP) at peak performance, typically expressed in picojoules per system operation (pJ/SOP).  & No & Yes & No & No & P \cite{chen2024epoc} \\ \hline
Power Density & Power delivery per area of the chip ($mW/cm^2$). Safe limits are imposed based on the functionality of the device. & No & Yes & Yes & No & P \cite{powerDensityExample} \\ \hline
\end{tabularx}
\end{table*}

\subsection{Accessibility}
We identify a large group of metrics (8 out of 13) that have high Accessibility. These metrics are, among others, the number of Parameters, Effective Synaptic operations, or Memory Accesses.
The main sources that mention or define these metrics are those works that focus on energy reporting and benchmarking~\cite{yik2025neurobenchframeworkbenchmarkingneuromorphic, analyticalsnn, hueber2024benchmarking}.

The reason for the high Accessibility of these metrics is that they can be obtained directly from the architecture of the model, or by running the model in non-neuromorphic hardware, while keeping track of the operations that take place.
There is no need to run the model on neuromorphic hardware at the development stages since the values reported should be the same regardless of the hardware.

In contrast, we observe another set of metrics that are not Accessible -- 5 out of 13. These metrics were obtained from literature that focuses on hardware design and testing \cite{ke2024neurobench, energyArea,chen2024epoc} and are common metrics to report results and compare performance to other architectures.

The low Accessibility of these metrics derives from the fact that they are obtained by directly measuring the hardware with power meters or other equipment, or through simulation tools.
Direct measurements by a model developer are usually not feasible because of the limited access to neuromorphic hardware.
For example, the SELF Lab only has one chip of this kind, and other popular neuromorphic architectures like Intel Loihi are only accessible through remote connection, making hardware measurements difficult to obtain.
On the other hand, according to the practitioners consulted, simulation tools in this field are usually closed-source, like Cadence Virtuoso -- a suite of tools for hardware design and simulation --, have a steep learning curve, and are hard to integrate with other tools, making their use hard for people not specialized in hardware design.

A common metric found across neuromorphic literature \cite{ke2024neurobench,chen2024epoc} is Energy per Learning and Energy per Inference, which represents the energy necessary, in Joules, to process a single training sample or inference.
These are useful metrics for comparing the performance and efficiency of different hardware architecture proposals.
However, these metrics, when measured on the hardware, do not give insights into the model itself, since hardware energy usage also depends on other components or power leakage from the architecture.

Other generic metrics for neuromorphic implantable devices that do not necessarily focus on AI performance are Peak per Energy Consumption, which represents energy consumption of individual system operations, or Energy-Area Figure of Merit (E-A FoM).
This last metric is relatively novel, proposed by Zhu et al.~\cite{energyArea} in the field of implantable chips.
This metric combines energy and chip sizes to introduce form factors into energy efficiency measurements.
However, this metric has still seen limited adoption.

\subsection{Fidelity}
When analyzing the metrics from the Fidelity perspective, we find that 8 out of 13 metrics present Low Fidelity.
We also see that this has an opposite relationship to Accessibility -- all of these metrics with Low Fidelity are previously identified as Accessible.

Metrics that have low fidelity come from software benchmarking of the SNN, and, for most of them, a proportional relationship to energy usage can be assumed.
However, deriving an accurate estimation of energy consumption is difficult, since it will depend on other parameters, like the target hardware architecture and the energy costs of operations in this hardware.

An example of a metric that presents this proportional relationship with energy is the number of Effective Synaptic Operations that appears proposed in various papers \cite{yik2025neurobenchframeworkbenchmarkingneuromorphic, analyticalsnn}.
This metric counts the number of MAC and AC operations taking place during inference in the connections between neurons, and it is similar to the floating point operations metric (FLOPs) used in Green AI literature as a proxy to energy consumption~\cite{flopsToEnergy}. 
However, this metric alone does not provide enough information to obtain an accurate estimate of energy consumption, since detailed information about the target hardware is required~\cite{flopsCriticism}.

On the other hand, the high-fidelity metrics we identified -- 5 out of 13 -- are also not Accessible.
These metrics are obtained through direct experimentation in the hardware with power meters or through detailed hardware simulations.
In the reviewed literature, examples of the latter are much more abundant.
For example, in the work by Yaldagard et al. \cite{rram}, a neuromorphic chip design is loaded into the Cadence Virtuoso Suite, given a set of tasks for benchmarking, and electric currents across the chip are simulated.

The energy measurements obtained from these methods accurately represent the energy consumption. 
Readings from a power meter directly represent energy consumption, while measurements from simulation tools are precise enough to be considered valid results for scientific literature, being used across the reviewed papers \cite{rram, energyArea}.

\subsection{Actionability}
Looking into the metrics that show the Actionability property, we observe that only two of the collected fulfill our definition of Actionable.
We identified these metrics as Actionable after our discussion with practitioners of the SELF Lab.

The first Actionable metric is Activation Sparsity, which represents the density of zeros across the network when performing inferences.
A higher sparsity means fewer calculations in the network, and efficiency is higher.
The reason to consider this metric as Actionable is that, in the field of SNN research, a common rule of thumb is that a model with sparsity lower than 60\% is considered an inefficient model that does not fully exploit the capabilities of neuromorphic computing.
The discussion with the practitioners from the SELF Lab confirmed this rule of thumb.
Therefore, if a developer observes an Activation Sparsity lower than 60\%, they know they must take action to improve the design.

The second identified Actionable metric is Power Density.
The reason for its Actionability is that health organizations, like the Food and Drug Administration (FDA) in the United States, set maximum safe limits on power delivery for implantable devices based on their functionality.
For example, for RF-emitting medical devices, the power density is limited to 10 mW/cm².
If the power density metric is above the limits set for the target application, the developer knows that they must change their design.

\subsection{Trend-Based}
While the majority of the identified metrics are not Actionable, 5 out of the 11 non-Actionable metrics are Trend-Based.
Trend-Based metrics can give a sense of how changes to the model might affect energy consumption during deployment, giving some Actionability if they are actively monitored during development.
The identified Trend-Based metrics are Effective Synaptic Operations, Membrane Updates, Activation Sparsity, Memory Accesses, and Training Time, which are also Accessible.

The reason these metrics are considered Trend-Based is that a proportional relationship to energy usage can be assumed, even if they cannot fully represent energy consumption for the whole model.

In a context of implantable devices, where we have a single, well-defined target architecture, we can safely assume that a higher number of Effective Synaptic Operations will translate into higher energy usage.
Keeping a record of these metrics across different model versions can help the developer assess how they compare in terms of energy efficiency.

\section{Discussion}\label{sec:discussion}

In this section, we identify two key challenges arising from our studies: the lack of metrics that are both accessible and high fidelity, and the lack of actionable metrics.
To address these issues, we provide recommendations for potential research directions.
Additionally, we discuss the threats to validity and limitations of our study.

\subsection{Bridging Accessibility and Fidelity}

The first challenge we can identify is the lack of metrics that are both Accessible and have High Fidelity.
This comes from the fact that literature on SNN benchmarking mostly focuses on the software side, measuring metrics that stem from SNN model design, and they fail to consider the impact of the target hardware
This is a highly relevant challenge, since the lack of Accessible metrics that can accurately reflect the energy consumption of an SNN model means that obtaining information about the energy efficiency of a model is not possible until a very late phase of the project.


A recommendation to solve this problem is to build energy estimation methods for SNN models, considering specifications about the chip provided by the hardware designers.
This estimation should not only account for the energy usage of the model, but also the energy usage of other features of the device, like pre- and postprocessing or wireless data transmission.

According to the practitioners we consulted, some of the specifications that could be provided from simulations are energy costs of MAC and AC operations, energy costs of memory accesses, or energy costs of processing in between layers, for mixed analog-digital neuromorphic designs.
Other specifications for features unrelated to the model could be the CMOS architecture used for chip fabrication, which determines standby power leakage.
For certain applications, other costs have to be considered. For example, for a brain implantable device, an Analog-Digital Converter is necessary to convert the brain waves readings to digital data for the model.

For this estimation to be useful, it should be divided into costs that come purely from the SNN model and costs that come from the other elements of the chip.
Depending on the target application for the hardware design, the difference in energy and power consumption of the model itself can be several orders of magnitude smaller than the power used by other systems \cite{epilepsyCnn}.
Therefore, in some cases, it might not be useful to keep improving model performance for a negligible impact on total energy consumption.


\subsection{Lack of Actionable Metrics}
A second challenge identified is the lack of actionable metrics across the literature.
This lack of actionable metrics impedes the developer from making informed decisions about the energy performance of the model.
Actionable metrics during model design can provide early feedback for SNN models in projects that practice hardware and software co-design.
This early feedback for software is highly relevant, since iterating over different software designs is much faster than iterating and improving the hardware architecture.
For example, in mixed analog-digital neuromorphic hardware, the analog part of the hardware is not programmable, and its architecture partly depends on the model.
Therefore, it is important to know as much information about the energy efficiency of the model as possible in the early phases of development.

To tackle this challenge, we present four recommendations about possible research directions for actionable.
These focus on assessing metrics with inherent Trend-Based properties from Sustainable Software Engineering, metrics that detect large changes in power requirements, metrics for early evaluation of battery life, and metrics that measure energy-performance tradeoff.

\autoref{tab:proposed} presents some of the metrics that could come out of these research directions, evaluated according to our classification methodology.
We update the Accessibility and High Fidelity field, and include a * if the metric is both Accessible and with High Fidelity, assuming that a reliable energy estimation is built as mentioned previously.

\begin{table*}[ht]
\centering
\caption{Overview of potential research directions to address challenges in existing literature}
\label{tab:proposed}
\begin{tabularx}{\textwidth}{|lXccccc|}
\hline
\textbf{Metric}               & \textbf{Description} & \textbf{Accessibility} & \textbf{High Fidelity} & \textbf{Actionability} & \textbf{Trend-Based} & \textbf{Reference} \\ \hline
Energy Delay Product          & Product of energy consumption and execution time & Yes*  & Yes*  & No  & No & NA \\ \hline
Speedup                       & Time\_new/Time\_old, change of execution time after changes & Yes  & Yes  & Yes  & Inherent & \cite{speedup} \\ \hline
Greenup                       & Energy\_new/Energy\_old, change of total energy after changes & Yes*  & Yes*  & Yes  & Inherent & \cite{speedup} \\ \hline
Powerup                       & Speedup/Greenup. If greater than 1, new code uses more power on average & Yes*  & Yes*  & Yes & Inherent & \cite{speedup} \\ \hline
Estimated battery life        & Expected battery life given energy per inference and battery capacity & Yes* & Yes*  & Yes & No  & NA \\ \hline
Inferences per battery cycle  & Inferences that can be done with a full charge. Opposite of the previous one & Yes* & Yes*  & Yes & No  & NA \\ \hline
Accuracy-Efficiency Tradeoff & Energy costs against accuracy gains between versions of the model & Yes* & Yes*  & Yes & Yes  & \cite{efficiencyRatio} \\ \hline

\textit{* assuming energy estimation}
\end{tabularx}
\end{table*}

\subsubsection{Extending Trend-Based metrics}
Our review uncovered that while many metrics provide information about the energy consumption of an SNN model, not many of them do so in an Actionable way.
However, Trend-Based metrics can provide certain Actionability when actively keeping track of them.

To tackle this problem, we propose incorporating some metrics from the Sustainable Software Engineering field that are inherently Trend-Based and can prove useful in the context of neuromorphic computing in general and implantable neuromorphic devices. 
Some of these metrics are derived from energy usage or execution time, which means they directly depend on the energy estimation previously mentioned.

Some useful metrics during model development are Speedup, Greenup, and Powerup \cite{speedup}, which can indicate improvements in performance and energy efficiency across the development of the model.
Therefore, they are Trend-Based by definition, in the sense that they are already derived from historical data.

\subsubsection{Large Changes in Power Requirements}
In the context of implantable neuromorphic devices, power delivery is highly limited by practical and medical reasons.
This is highlighted by metrics like power density, for which health organizations establish safe maximum limits.
Therefore, metrics that easily indicate large changes in power requirements for the model are crucial for the developer.

The previously mentioned Powerup or the Energy Delay Product metric can be applied in this context.
These metrics combine execution time and energy to give a sense of total power usage.
This is extremely relevant for implantable brain devices, due to the power per area limits set for these applications. High power delivery and/or long execution times of the model can lead to elevated temperatures, which can be uncomfortable or damaging to the patient.

\subsubsection{Battery related metrics}
Another aspect of the model that cannot be quickly evaluated from the reviewed metrics is the estimated battery life of the device.
This is a highly relevant problem in the field of implantable brain devices, since batteries are implanted together with the chip through brain surgery, and cannot be recharged or replaced without further surgery.
Therefore, according to practitioners, a common quality attribute used in this field is that batteries should last for at least 10 years, to minimize the number of invasive brain surgeries performed as much as possible.

To cover this aspect of SNN development for implantable devices, we recommend research directions for defining and estimating metrics like Expected Battery Life or Expected Number of Inferences per Battery Cycle.
These metrics can be derived from an energy estimation based on specifications, battery capacity, and consumption of other components.
With them, developers can access an easy overview of the impact of the model on battery life, and making decisions that target energy efficiency is easier.

\subsubsection{Assessing Energy-Accuracy Tradeoff}
Another relevant aspect of model development that lacks Actionable metrics for evaluation is assessing the tradeoff between energy and accuracy of the model.
The traditional development pipeline of an AI model focuses on iterating designs to maximize accuracy.
However, sometimes small improvements in accuracy can come at a high energy cost.
In the implantable field, where power and energy are highly limited, sacrificing accuracy slightly could be worth it compared to the energy efficiency gains.

Possible research directions to solve this problem could be defining a metric or set of metrics that accurately represent the Energy-Accuracy tradeoff between model versions. 
This metric would be useful when iterating over different versions of the model.
Estimations of energy combined with the reported accuracy of the model can be used to report an increase in energy consumption against gained accuracy and help determine if potential accuracy increases are worth the additional energy cost.

A previous work by Cueto-Mendoza and Kelleher \cite{efficiencyRatio} defines a similar Efficiency Ratio metric for AI models, defined as $Accuracy/Energy$.
This definition is limited, since a model with good accuracy and a model with poor accuracy could show similar values if the second one uses much less energy.

Therefore, it is not a very actionable metric.
As an example, suppose we have a version V1 of a model that reaches 0.7 accuracy with a certain energy cost per inference, and a version V2 that reaches 0.8 using double the amount of energy.
Efficiency Ratio, as defined, would not give much insight into this breakdown.
A more sophisticated metric would help determine if it is worth doubling energy consumption in exchange for this additional 0.1 accuracy.
In the implantable context, it could even be combined with the previously mentioned battery-related metrics to determine if the hit to battery life would be significant.




\subsection{Threats to Validity}


The nature of this study presents some limitations.
Firstly, we focus our scope on SNNs, and specifically on implantable healthcare devices.
Therefore, our findings might not generalize to other domains of AI.
In general, we believe most of our reflections could be applied to the field of edge AI, given the similarities between both fields.
Edge AI also deals with low-power devices and sometimes with devices running on a battery.
Additionally, some reflections could be applied to the field of Green AI in general, since these metrics focus on reporting energy consumption.
However, the definition and validity of these metrics for other fields or applications would have to be evaluated independently.

The lens of the project through which we study the problem also limits the feedback from the practitioners.
These practitioners come from a concrete neuromorphic project from the SELF Lab.
Their views are shaped by this project, and some conclusions might not apply to other projects.

The limited research done in SNN benchmarking is also a limiting factor for the collection methodology applied for this review.
The research and first attempts at benchmark standardization started as recently as 2023.
Therefore, there is little literature to apply a systematic method, so we performed an ad hoc literature review. To mitigate problems derived from this method, we applied snowballing over the Neurobench paper \cite{yik2025neurobenchframeworkbenchmarkingneuromorphic} to cover a larger number of relevant papers.
Additionally, there could be new research that tackles these problems in the near future.

\section{Conclusion}\label{sec:conclusion}
Neuromorphic computing and SNNs are a promising new paradigm for improving the energy efficiency of Artificial Intelligence models.
However, as highlighted in this study, the research has focused mainly on evaluating hardware and models in conjunction, while research into only model benchmarking is relatively new.

By looking into the works that focus on SNN benchmarking, we uncovered and categorized the most common metrics for assessing the energy consumption of an SNN model, based on Accessibility, Fidelity, and Actionability
With this classification, we identified which metrics approximate energy consumption better and are easier to measure during model development.

We find out that there are no energy metrics that are both Accessible and present High Fidelity.
Benchmarking frameworks for SNNs focus mainly on model design and metrics, without considering the effects of target hardware.
This makes early evaluation of the model difficult, and we recommend research directions towards building reliable estimations that bridge Accessibility and Fidelity.

The study also highlighted the lack of actionable metrics for developers. Despite the variety of metrics, none of them focused on giving the developer early and approachable feedback.
Therefore, from the lenses of implantable neuromorphic devices, we proposed research directions to extend the catalog with actionable metrics, focusing on accurately and transparently representing energy and power trends during development, estimated battery life, and Energy-Accuracy tradeoff.

By identifying gaps in existing energy metrics for SNNs in neuromorphic implants and proposing research directions to bridge them, this study lays the groundwork for more effective benchmarking methods. 
Our recommendations open future research into energy metrics that provide quicker feedback and better Actionability, ultimately supporting developers in designing more efficient and practical SNN models.

\section*{Acknowledgments}

Our research is partially funded and aimed to support the work by the Smart Edge Lab for Healthcare (SELF Lab) from TU Delft.\footnote{\url{https://www.tudelft.nl/ai/self-lab}} 

\bibliography{IEEEabrv,references-4}
\end{document}